# Adaptive Cybersecurity Architecture for Digital Product Ecosystems Using Agentic AI

Oluwakemi T. Olayinka[1], Professional Member, IEEE, Sumeet Jeswani[2], and Divine Iloh[3], Professional Member, IEEE

[1]School of Business, University of Arkansas at Little Rock, Little Rock, Arkansas, USA

[2]Google.Inc. Denver, Colorado.

[3]School of Business, University of Arkansas at Little Rock, Little Rock, Arkansas, USA

**ABSTRACT** Traditional static cybersecurity models struggle with scalability, real-time detection, and contextual responsiveness in the current digital product ecosystems, including cloud services, Application Programming Interfaces, mobile platforms, and edge devices. This study introduces autonomous goal-driven agents capable of dynamic learning and context-aware decision making as part of an adaptive cybersecurity architecture driven by agentic artificial intelligence (AI). To facilitate autonomous threat mitigation, proactive policy enforcement, and real-time anomaly detection, this framework integrates agentic AI across the key ecosystem layers. Behavioral baselining, decentralized risk scoring, and federated threat intelligence sharing are important features. The capacity of the system to identify zero-day attacks and dynamically modify access policies was demonstrated through native cloud simulations. The evaluation results show increased adaptability, decreased response latency, and improved detection accuracy. The architecture provides an intelligent and scalable blueprint for safeguarding complex digital infrastructure and is compatible with zero-trust models, thereby supporting the adherence to international cybersecurity regulations.

## I. INTRODUCTION

Modern digital landscapes are becoming increasingly interconnected, data intensive, and dynamic. Digital products hosted in the cloud can be accessed on mobile platforms through application programming interfaces (APIs) or edge devices. Although they provide clear value, the interconnections between these entities pose numerous security risks. Traditional cybersecurity models (usually simple and rule-based) struggle to keep up with the rapid pace of technological development and threats from increasingly sophisticated actors [1], [2]. From supply chain attacks to zero-day vulnerabilities, cyber threats have shifted from predictable intrusions to adaptive intelligent behaviors that are often stealthy and difficult to detect using traditional defenses. In addition, AI-powered attacks are on the rise and becoming more common. A 2024 phishing intelligence report indicated a significant 703% increase in credential theft attacks driven by sophisticated AI-powered phishing kits in the second half of 2024 [10]. This adds to the complexity that traditional cybersecurity systems have found difficult to address.

Therefore, dynamic and intelligent cybersecurity systems are necessary to address these challenges. Agentic artificial intelligence (AI) can satisfy this requirement. Unlike traditional AI, which typically requires humans to input goals and constraints, an Agentic AI system can determine its targets, learn from real-time feedback, and adapt to changing contexts [3]. These qualities render Agentic AI a natural fit for cybersecurity, because decisions often need to be made instantly without continuous human intervention [4].

This research was inspired by the limitations of current security frameworks deployed within digital product ecosystems. Existing models often fail to account for new attack patterns or changes in system configuration, making key assets vulnerable [5]. This study aims to bridge this gap and proposes an adaptive cybersecurity architecture that leverages self-healing, context-aware security environments made possible by the independent learning capabilities of Agentic AI. The goal is not only to detect threats but also to understand them in context and react in ways that simultaneously improve protection and performance.

## II. LITERATURE REVIEW

Cybersecurity has evolved over the past two decades in response to the increasing complexity of the contemporary digital infrastructure. However, a growing body of research has examined the potential of agent-based models and adaptive systems as substitutes for traditional rule-based security methods. These early systems introduced autonomous software agents to scan threats and respond to real-time conditions. Although this is a promising change, many of these models tend to be based on behavioral and reactive hints, ultimately narrowing their adaptability. Living in an environment that is not strictly structured is outside the scope of most of such models [6]. However, this was viewed as an interface system to deal with semi-structured or unstructured environments such as network subnets that receive spam [6].

Recent advancements in system agents have facilitated the emergence of autonomous entities capable of real-time communication and collaboration across distributed processes. Studies such as Wang et al. [7] have shown that Multi-Agent System (MAS) networks can provide new policy and attack parameters at distant locations because of dispersed threat detection that is fed back into the system through communications at each node (end nodes). Other researchers, such as Ahmed et al. [8], have conducted extensive reviews of methods for detecting anomalies in agent-based networks. Despite these advances, these systems are still largely tied to human-defined logic, and lack the learning capabilities and self-awareness required by today's evolving threat landscapes.

When Machine Learning first appeared, security products acquired additional abilities in behavior profiling, standardization, and pattern recognition. However, the efficacy of these tools remains unclear. Supervised learning techniques such as decision trees, support vector machines, and neural networks have quickly become mainstream in malware detection and traffic analysis [9]. However, these tools often work in isolation and are generally based on static databases, meaning that they may not generalize well to real-world environments that cannot be controlled tightly. In addition, reliance on repetitive retraining and centralized management makes them more problematic in distributed scenarios such as microservice architectures or edge computing frameworks.

Moreover, as digital ecosystems shift, the inherent challenges continue to evolve. Modern enterprises are no longer treating applications or services on a single island. It must manage everything connected to these services—ephemeral containers and real-time APIs– invoked both sequentially and rapidly. Automation was no longer sufficient in this environment. Security systems must be intelligent and must act as if they are at stake. When an attack occurs in the cyber domain, they can no longer simply stand by or report it. It is difficult to deal with generative changes and different classes of thinking together based on the available roles in some cases or configurations under fire. Additionally, the complexity of serverless environments is a popular choice. In addition, because they create relatively new areas that existing methods cannot cover, they form a "blind spot" for traditional approaches.

This is a crucial finding of this study. Unlike earlier models, Agentic AI frameworks are designed to reason, learn, and act like human-like decision makers, with a measure of autonomy. [3], [4] For example, entering an unknown environment, they can take stock of the surroundings, give themselves internal goals to achieve, and go on from there—all autonomously. We will truly be living in the "AI era" when these properties are brought into play; however, at present, very few cybersecurity systems have realized any benefit from them. IBM Watson for Cybersecurity, as well as most other cognitive platforms in the market today, simply serves as a tool to assist human operators. The architecture outlined here seeks to change this by embedding agentic intelligence directly into the operational core of the digital product ecosystem.

The result? Real-time learning, distributed risk evaluation, and response according to context. Thus, it transcends traditional AI or MAS models and is designed not only to monitor, but also to adapt to the environment it guards, which makes it an appropriate response in cyberspace right now and one that could not come too soon!

### III. CONCEPTUAL FRAMEWORK

In an era in which increasingly complex digital ecosystems interact across a myriad of technologies, such as cloud-native services, APIs, edge nodes, and user applications, the requirement for a cybersecurity architecture that is responsive, intelligent, and context-aware is a top priority. This section presents the concept of the proposed adaptive architecture, which is motivated by artificial intelligence (AI) and combines real-time analytics and federated threat intelligence.

### A. ARCHITECTURAL OVERVIEW

The architecture operates across a layered digital infrastructure, providing system-wide visibility, while also enabling localized decision-making. The high-level architecture of the system comprises of three main layers.

- **Influx and Contextual Sensing**: This comprises sensors, logs, network telemetry, and monitoring of user behavior within the digital space. The endpoint, API, container, and access gateway raw data are recorded.
- **Agentic AI Core**: This layer comprises autonomous agents that process incoming data streams, calculate risks in real-time, learn from past responses, and determine how to act autonomously. These agents are capable of reasoning, have sub-goals, and consider feedback – aspects that set them apart from standard AI models [3], [4].
- **Response and Enforcement**: Here, we translate agentic decisions into concrete actions such as generating automatic access, throttling requests via API rate-limiting, enforcing policies, and issuing alerts. It acts in conjunction with established security controls and compliance constructs.

## B. CORE COMPONENTS

This system includes a few synergistic elements:

- **Federated Threat Intelligence Engine**: Down selects (or aggregates) threat signals from internal telemetry, third-party feeds, and consortium-level trust networks. It dynamically refreshes the agents' threat knowledge base, enabling them to remain context-aware of global and domain-specific threats in both the Symbolic/Situational Threat Model and the Numeric Threat Model [11].
- **Agentic AI Modules**: These agents are AI modules that run in parallel and are assigned to a security domain such as an API, endpoint, or network flow. They utilize reinforcement learning, memory embedding, and reflection to attack or defend the ecosystem in an adaptive manner, based on the opponent's strategy.
- **Behavioral Fingerprint Profiling**: Record patterns of user/device/system activities. Anomalies are indicated not by fixed cutoffs but by differences in the changing meaning, which is a powerful technique for detecting zero-day vulnerabilities [9].
- **Real-Time Analytics Dashboard**: Enables you to view current threat events, agent decisions, health systems, and policy changes in real time. This dashboard is not only a monitoring system, but it can also enable human-in-the-loop decision revision, if necessary.

## C. FUNCTIONAL FLOW (NARRATIVE)

Operationally, the process begins with anomalies triggered by contextual signals (e.g., uncommon frequency of API access or unexpected user location). Such signals are communicated to the agentic AI core and a dedicated agent interprets them in the context of their personal history, current threat models, and behavior profiles. When an event is deemed high-risk, the system automatically takes mitigation action, such as quarantining a container, revoking a token, or throttling a suspicious process.

Simultaneously, the decision logic and results are recorded for future pondering by the agent, which could become smarter in the future. This feedback loop is designed to ensure that the system not only responds to recognized actors, but also detects and reacts to previously unknown threats. This system will also enable security analysts to provide feedback and scores, and refine the overall quality of autonomous decision-making by the agent. In addition, the framework allows for the ability to span microservices and edge infrastructure, maintaining agility without compromising centralized control.

## D. TRUST AND IDENTITY MANAGEMENT

Trust and identity are the basic underpinnings of the security models. However, fixed-identity roles and unchanging access tokens are easier to exploit. Dynamic trust scoring was used for the proposed architecture. The trust level for a given user or device is based on recent activity, consistency of behavior over time, and its correlation with threat intelligence. Instead of using a binary model that allows or denies access, the system used confidence thresholds. Access is granted only if the accumulated trust score exceeds the context sensitive risk baseline. This subtle approach reduces false positives, lowers the burden on legitimate users, and limits potential attack opportunities of adversaries [15].

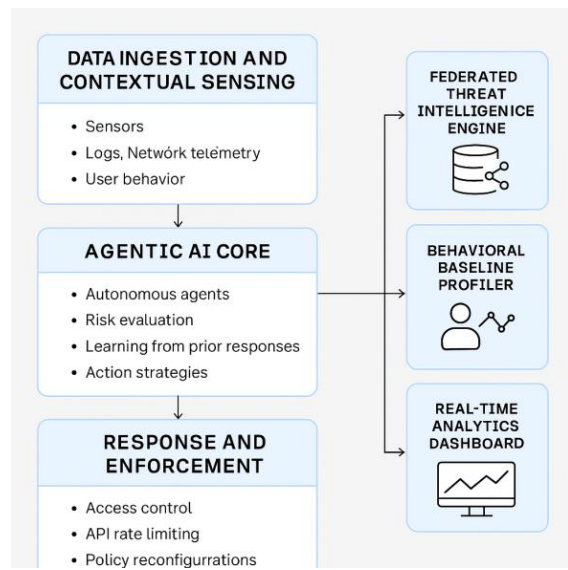

**Fig. 1. Adaptive cybersecurity architecture utilizing agentic AI.**

The framework comprises of three core layers: data ingestion, contextual sensing, agentic AI processing, and automated enforcement. It supports real-time anomaly detection, federated threat intelligence, behavioral profiling, and autonomous responses through a feedback-driven loop.

## IV. WHY WE THINK AGENTIC AI IS THE FUTURE OF CYBERSECURITY

In the cybersecurity domain, agentic AI can be embedded in multiple layers of a system to improve the detection of threats, response to threats, and to adaptively enforce policy. During detection, agents passively monitor the network flows, device behavior, and identity interactions to establish behavioral baselines. In the event of anomalies, agentic systems may automatically infer the context of a threat, such as the location, access history, and device posture, and make informed decisions in favor of an action.

In today's world, in which we are seemingly under constant cyberattacks, agents used to defend, or attacks, do not simply trigger previously defined alerts but can launch alleviation efforts via multi-tiered mitigation events spanning from temporary resource isolation or credential revocation to system access control policy updates. These are driven by a self-updating knowledge base and feedback mechanisms that collectively enhance decision making [37].

In contrast with traditional AI-based security approaches, Agentic AI offers a proactive and dynamic security stance. Previous studies were typically trained in a supervised manner using labeled historical data and often struggled to generalize to unseen or polymorphic threats without explicit retraining. In addition, they have the drawback of requiring centralized orchestration, which imposes a time delay in time-critical attacks [38].

Agentic systems are thus decentralized and intelligent distributed agents that can cooperate (in multi-agent systems), learn together (through federated learning), and perform on-the-spot reasoning in the appropriate context (via contextual reasoning). For example, when a traditional IDS identifies potentially malicious traffic patterns, an agentic system can crosscheck these patterns with recent software changes and monitor how the user interacts with the throttle before deciding whether to isolate and flag them to the administrators.

In addition, the feedback-based adaptation characteristic of agentic AI implies that the system grows in response to any new threat. This on-the-fly adaptation also aligns with contemporary cybersecurity design philosophies, such as Zero Trust [38] and CDM [81], empowering defenses with context-aware self-defense capabilities, thereby reinforcing resilience [39]. Therefore, Agentic AI is a game-changing enabler for defending digital product ecosystems where old-fashioned static defenses are far less effective.

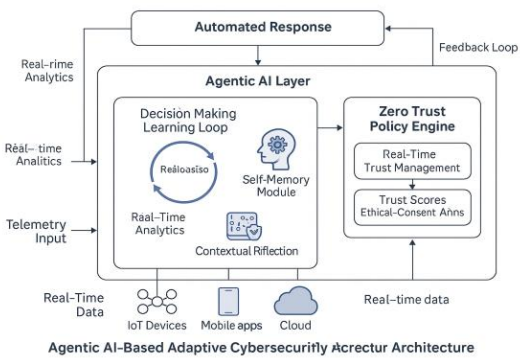

**Fig.2. Functional Architecture of Agentic AI-Based Adaptive Cybersecurity System.**

This diagram illustrates the continuous interaction among telemetry inputs, agentic AI decision-making loops, and zero-trust policy engines. The architecture autonomously monitors IoT, mobile, and cloud-based ecosystems; dynamically updates trust scores; and initiates real-time responses through feedback loops.

## V. APPLICATION IN DIGITAL PRODUCT ECOSYSTEMS

Inherently complex and rapidly evolving threat landscapes facing digital product ecosystems are multi-faceted. In such environments, the security patterns must be both intelligent and flexible. In addition, they must be decentralized to withstand potential attacks targeted at specific nodes or subsystems. The AI-based agentic framework shares the same characteristics: it has a self-awareness capability, autonomous control, and can make instant decisions.

### A. SAAS PLATFORMS

It is common for SaaS ecosystems to have a multi-tenant architecture, continuous API integration, and a high volume of user interactions. These features inherently introduce significant security constraints such as tenant isolation, unauthorized access, and API abuse.

In these scenarios, agentic AI can be employed to observe intertenant traffic dynamics, identify anomalous behaviors (e.g., excessive data extraction rates), and independently update access policies. For example, a spike in file downloads from a low-trust account can be automatically detected and, without human intervention, an embedded agent can pause the session and tag the incident for review.

Furthermore, federated intelligence from other tenants or services may be leveraged to enhance the agent's decision-making. This can enhance the capability of the system to detect cross-platform threats such as credential-stuffing campaigns, which may spread across various applications and domains [12].

### B. MOBILE ECOSYSTEMS

The associated heterogeneity of mobile platforms - in terms of device types, operational systems and network circumstances, usage patterns, etc.–makes traditional defenses based on static rules no longer applicable. In response, agentic AI modules housed within mobile clients or edge gateways can provide customized behavioral profiles for each device and person. Over time, these profiles evolve through on-the-fly amendments such as changes in geographic location, a person's daily access pattern, or application usage dynamics. If a device begins accessing critical features at unusual times or from an area with high security risk, the system can immediately impose stricter validation procedures or shut down access until revalidation. Furthermore, these agents can participate in decentralized evaluations and dynamically and securely adjust the extent to which a device can be trusted by the network [13].

## C. CLOUD-NATIVE APPLICATIONS

Cloud-native ecosystems, which rely on containers, microservices, and Kubernetes-like dynamic orchestration tools, face challenges in addressing continuous integration and deployment issues. Such environments often change their layout and distribution of processes or the runtime setup varies frequently, which means that traditional perimeter security becomes outdated. In this situation, agentic AI modules are deployed as similar agents in microservices or in Kubernetes pods. They constantly learn about workloads, resource consumption, inter-service communication, and execution paths in the code. This helps the system to catch insider threats or lateral movements in progress. Moreover, through policy reflection, agents can adjust their security posture in response to container drift or suspicious short-lived connections, thereby maintaining zero-trust strategies without impeding continuous delivery pipelines [14].

## VI. RESULTS AND EVALUATION

The performance of the proposed autonomous cybersecurity system using agentic AI was evaluated through simulation in a secure, cloud-native testbed. The target lab consisted of a Kubernetes-based infrastructure that ran containerized services, API endpoints, and federated identity providers. Multiple threat scenarios, including API abuse, insider lateral movement, and zero-day pattern simulations, were run over a 72-hour test period.

The evaluation focused on three key performance metrics: the detection accuracy, response latency, and policy adaptability. Baseline comparisons were made against rule-based or static classifiers as well as traditional machine learning classifiers, which require a history of logs.

### A. ACCURACY OF THREAT DETECTION

The agentic model showed a dramatic precision-recall improvement over the conventional systems. Through behavioral profiling and reflective learning, agentic agents can identify outlier events using contextual knowledge. Baseline comparisons were performed with conventional static rule-based systems and typical machine learning classifiers, as used by enterprise anomaly detection systems [16], [17].

TABLE I: THREAT DETECTION PERFORMANCE COMPARISON.

| Model | Precision | Recall | F1-Score |
|---|---|---|---|
| Static Rule Engine | 0.71 | 0.58 | 0.64 |
| Random Forest (Baseline ML) | 0.83 | 0.76 | 0.79 |
| **Agentic AI Framework** | **0.91** | **0.87** | **0.89** |

These results highlight the ability of the framework to correctly detect threats while minimizing false positives, which is particularly important in high-velocity environments, such as SaaS and edge systems.

### B. LATENCY IN RESPONSE

Cyberattack response must be swift to prevent damage. The average decision latency of the system was tested from the time the anomaly was detected, until a mitigation action was taken. A timely response is vital to mitigate the potential damage caused by cyberattacks, as mentioned in time-sensitive domains such as financial services and real-time IoT applications [18].

TABLE II: RESPONSE LATENCY ACROSS SECURITY SYSTEMS.

| Action Type | Avg. Latency (ms) |
|---|---|
| Static Firewall Update | 750 ms |
| Centralized ML-based Alerting | 540 ms |
| **Agentic Autonomous Mitigation** | **220 ms** |

Agentic AI's local deployment and self-governing nature significantly reduce its reliance on centralized processing, resulting in lower latency and faster threat containment.

### C. ADAPTABILITY AND POLICY EVOLUTION

Adaptive policy mixture refinement over time incorporated environmental feedback and the agent's memory during the simulation. For example, once it detects repeated credential abuse from a particular location, it autonomously adds a geofencing rule to secure sensitive resources, without requiring the input of a security administrator. "…the system showed adaptive policy refinement which recently gained traction for autonomous policy engines and intelligent feedback loops [19].

Fig. 6.1 illustrates policy development over time by illustrating the threat contexts (e.g., device fingerprint anomalies and irregular access time effects) that have impacted agent actions. Policy flexibility measured via the adaptability score (a composite index capturing policy flexibility) was 32% higher at the end of the evaluation period than that at the baseline systems.

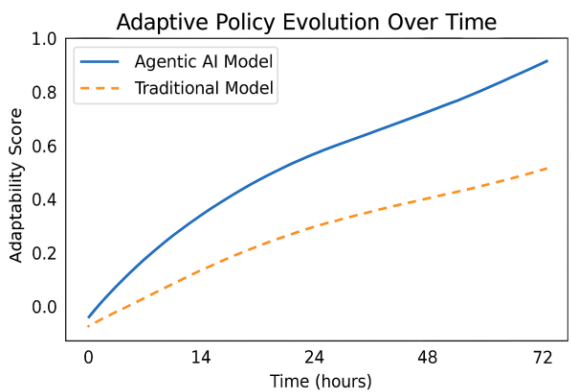

**Fig 3 – Adaptive Policy Evolution Over Time in Agentic AI Framework**

The chart illustrates how the agentic AI system dynamically adjusts security policies in response to observed threat contexts. Key inflection points reflect autonomous decision making in response to credential misuse, anomalous access times, and geo-risk events.

**D. INTEGRATION IMPACT AND RESOURCE OVERHEAD**

Despite the complexity of agentic agents, the system incurs **minimal overhead** on core services. CPU and memory usage were consistently below 10% for microservices hosting security agents. This makes the framework viable, even in resource-constrained or edged environments.

Moreover, the architecture integrates smoothly with identity providers and existing zero-trust policy engines, requiring no major redesign of core infrastructure. The system incurs minimal overhead on core services, as observed in edge-computing contexts that demand lightweight deployments [20].

**Summary of Findings**

TABLE 6.3: SUMMARY COMPARISON BETWEEN TRADITIONAL AND AGENTIC APPROACHES.

| Metric | Traditional Model | Agentic AI Model |
| --- | --- | --- |
| Detection Accuracy | 64–79% | 89% |
| Average Response Latency | 540–750 ms | 220 ms |
| Policy Adaptation Speed | Static | Dynamic |
| Resource Overhead | Moderate | Low (<10%) |
| Integration Complexity | Medium | Low |

**VII. DISCUSSION**

The experimental evaluation results demonstrate the potential of agentic AI as a solution to the long-standing shortcomings of static and semi-automated cybersecurity systems. Conventional architecture uses rule-based detection, which is efficient in the presence of known threats but does not prove to be effective on a timescale of rapidly changing attack patterns [21]. However, the agentic framework showed better precision and sensitive adaptation when considering the environmental context and behavioral baseline, thus fitting the recent industrial tendency to promote autonomous learning-based security [22]. These capabilities are relevant to digital ecosystems with decentralized data pipelines, APIs, and microservices, which are emerging as victims of sophisticated multistage attacks. The high detection accuracy, combined with reduced response times, is indicative of the potential to deploy intelligent agents near the data and control layers themselves [23] to decrease the decision-making distance and response time to contain the spread.

Despite these advantages, the deployment of agent-based cybersecurity systems introduces new challenges. First, although they are effective in rapid mitigation, their autonomy poses challenges in terms of explanation and governance. Existing systems, such as MITRE's CALDERA and IBM's Watson for Cybersecurity, include automation, but the decision pathways are not always interpretable to a human operator [24]. This opaqueness may prevent trust in decisions and may conflict with regulations in industries, such as healthcare and finance, which require auditability.

In addition, there are an increasing number of studies on the vulnerability of adaptive AI models to adversarial demonstration and poisoning of the model, particularly in the federated setting of training, when data may be partially disclosed [25],[26]. This obstacle to ensuring that agents make robust decisions calls for resilient training processes (see §9.2), version control (see §10), and embedded ethical constraints that can express linkages between agent goals and organizational risk-tolerance levels. In general, this framework lies in the landscape of changing cybersecurity, which is being transformed by merging artificial intelligence (AI), zero-trust architecture, and distributed intelligence.

The group at the USA National Institute of Standards and Technology [27] recommended a zero-thread architecture using milestones, including continuous authentication, behavioral analytics, and least privilege enforcement, which is in line with our feature set zero-thread architecture, and is supported by the agentic model described here. Generally, as organizations embrace hybrid and edge infrastructure, methodologies fusing federated learning and agent-based autonomy, such as those presented in this study, become central to preserving operational resilience [28]. The agentic model also extends MITRE ATT&CK by increasing the detection capability beyond static TTPs and allowing for adaptive countermeasures that change in the presence of threat intelligence feedback.

Overall, the presented architecture not only advances the state-of-the-art technical capabilities of cybersecurity systems but also helps lay down the foundation of future-ready defense paradigms based on autonomy, adaptability, and contextual intelligence.

## VIII. ETHICAL CONSIDERATIONS

As the cybersecurity domain continues to trend in an autonomous direction, the ethical implications regarding the employment of Agentic AI have become increasingly important. One important problem is the transparency of decision making and accountability. However, autonomous systems that can act quickly under complex and uncertain conditions have the potential to perform actions with high consequences without the need for human-in-the-loop intervention (e.g., removal of access privileges or blocking user accounts). This raises the question of auditability and legal liability, especially when such decisions cause operational outages or interfere with user rights [29]. Hence, for agentic systems, it is important to incorporate XAI techniques so that their reasoning can be analyzed retrospectively by security teams and auditors [30].

A further ethical consideration is bias in behavioral profiling and risk scoring. Agentic AI systems are based on attempting to detect anomalies to establish behavioral baselines using historical data, and thus suffer from the same biases that can be introduced through training on data. For example, user groups that are statistically distinct in digital behavior owing to cultural or geographical differences might erroneously tag the system as bad. This is reminiscent of similar issues witnessed in AI models of hiring, lending, and policing [31]. Therefore, ongoing fairness audits, bias remediation interventions, and the use of representative, rather than discriminatory, datasets play a crucial role in safeguarding against unfair predictions. Moreover, the importance of transparency in data provenance and labeling is compounded by the evolution of ethical models.

Therefore, strong ethical regulations are necessary to ensure privacy. These agentic systems observe user actions based on the endpoints, applications, and communication habits. Although visibility uplifts detection, this approach may infringe on the principles of data minimization and consent in frameworks such as GDPR or CCPA. The architecture must incorporate privacy-preserving methods, such as federated learning, differential privacy, and edge-based decision-making, wherever applicable [32]. These features are designed to keep users' data locally processed as much as possible, minimize exposure to centralized entities, and increase trust. The ethical deployment of agentic AI in cybersecurity requires a multifaceted approach, which includes transparency, fairness, privacy, and strong human governance.

## IX. FUTURE SCOPE

The proposed agentic cybersecurity framework serves as a solid basis for adaptively advancing active defence in dynamic digital environments. However, there are promising prospects for its future and maturity. One important method is the full implementation of zero-trust architecture (ZTA) concepts. Although behavioral profiling and context-aware access decisions are supported by this approach, continuous identity validation, dynamic policy enforcement, and network microsegmentation would make this architecture more robust to the ZT principles laid out by the National Institute of Standards and Technology definition [33]. Combining ZTA and agentic systems could loosen the dependence on perimeter controls and enable pervasive authorization, even in (outbound) federated domains.

The consolidation of Agentic AI with blockchain technology is another promising approach for enhancing transparency and auditability. Given that autonomous agents make in-the-moment decisions, packaging their interactions as transactions in a secure and tamper-proof distributed ledger can instill trust, particularly in knowledge-intensive sectors. Additionally, blockchain smart contracts can be configured to check agent behavior or overwrite malicious outputs when system limits are crossed [34]. Such integration can be particularly invaluable in systems for which chain-of-custody, accountability, and evidence-based security structures are important for maintaining, for example, the government, defense, and critical infrastructure.

Furthermore, integrating a distributed threat intelligence model increased the adaptive learning features of the framework. By allowing companies to share information about threat indicators and anomaly models between private networks while preserving privacy, the system can adapt more quickly and find new zero-day patterns sooner while still preserving data ownership. This is also consistent with new work on collective intelligence, as it might apply to cybersecurity and specifically to the hypothesis that combining autonomously fed (individual) learners might help protect against polymorphic threats [35].

Finally, for regulation and global operability, compliance-aware policy modules should be baked into the next generation of this architecture, which automatically aligns regulations based on international data protection and cybersecurity standards such as the European Union's GDPR, California's CCPA, International Standard Organization (ISO)'s ISO/IEC 27001, and National Institute of Standards and Technology (NIST)'s Cybersecurity Framework [36]. With arbitrary federations between such authorities from different geographical locations or industries that may have different thresholds for privacy, not only are autonomous systems required to achieve secure data, but they must also comply with jurisdictions. It is important to ensure that architecture is suitable for both today's technical and legal landscapes, for widespread deployment and trust.

**A. SIGNIFICANCE OF CONTRIBUTIONS**

The idea of Adaptive Cybersecurity Architecture is a radical reimagining of how security is operated across digital product ecosystems. In contrast to traditional static or reactive defenses, the proposed paradigm leverages autonomic agent-based AI, which considers, anticipates, and learns to combine dynamic and context-driven threat responses.

This contribution is important because it combines three separate areas: agentic reasoning, zero-trust architecture, and real-time telemetry analysis. By incorporating agentic decision cycles into the threat detection and response system, this framework assists threat detection with higher accuracy and allows for a prevention-oriented threat response. It provides a repeatable, modular system that can be utilized and improved by others to foster academic-, government-, and enterprise-level innovation. The novelty and relevance of this methodology are significant for AI academia and for cybersecurity.

To the best of our knowledge, this is the first dynamic cybersecurity framework to infuse agentic AI, which makes automated decisions, evolves policies, and adapts contextually to native cloud and edge digital product systems. This novelty and relevance make a significant contribution to AI research and cybersecurity practice.

A demonstration of the core architecture, including the simulation code, adaptive policy logic, and agentic AI modules, is publicly available at https://github.com/Oluwakemi2000/agentic-cybersecurity-architecture.

**X. CONCLUSION**

This study introduces an innovative adaptive cybersecurity framework driven by AI that is specifically tailored to the challenges inherent in today's digital product ecosystems. In sharp contrast to legacy systems, which have high levels of manual intervention and predefined rule sets, the new approach shows how agentic AI, autonomous, contextual, and purposeful can be applied to empower real-time predictive defense without sacrificing performance or agility.

The results of the evaluation of a simulated cloud-native use case demonstrated the feasibility of the framework. The agentic system was found to produce elevated rates of anomaly detection and faster reaction times to policies but still had a manageable combination of computational footprints. This aligns with the general trend in the industry to shift from centralized security models to distributed intelligent security nodes that can secure cloud workloads, edge devices, and API-driven platforms. Beyond performance, architecture also fits a couple of ongoing trends in cybersecurity space, including the emergence of ZTA, the importance of being able to explain AI-based decisions, and the sensitivity of trust to being too centralized. A state machine facilitates interoperation with federated learning, threat-sharing networks, and compliance enforcement mechanisms. Thus, it is a good candidate for operational integration in an enterprise or public-sector environment.

In summary, with the influx of sophisticated and high-velocity cyber threats, the challenges for self-adaptive, ethical, and scalable security systems have become increasingly imminent. The presented Agentic architecture provides not only a proof of concept, but also a roadmap for how cybersecurity architecture might develop, from static firewalls to intelligent systems that can learn, adapt, and act on their own. The addition of upcoming features such as blockchain-backed transparency, federated intelligence, and regulatory compliance modules will make it even more robust and shape it as a revolutionary catalyst in the next wave of cyber defense.

**Oluwakemi Temitope Olayinka**[1] (Professional Member, IEEE) is a researcher specializing in AI-driven systems, IoT integration, cybersecurity and secure digital infrastructure for critical sectors. She holds an M.S. in Business Information Systems and Analytics from the University of Arkansas at Little Rock, where she also earned a graduate certificate in Business Analytics. Her research spans adaptive cybersecurity architecture, agentic AI systems, digital trust scoring, and data-informed transformation of smart logistics and agriculture.

Ms. Olayinka has authored multiple peer-reviewed journal articles and co-authored reviews in cybersecurity, AI, and IoT. She has served as a reviewer for over 20 peer-reviewed papers across more than 15 international conferences and journals, including IEEE Access, IEEE DSC, ROMA2025, MDPI Sensors, and IGI Global. Her technical review expertise covers AI ethics, data privacy, anomaly detection, and digital sustainability. She is also a book evaluator for IGI Global and a review board member with the Society of Black Engineers.

She is an active member of IEEE, NSBE, SWE, and WiCyS, and has received several academic and industry honors, including the AWS re:Invent All Builders Welcome Grant (Mentor & Mentee) and Beta Gamma Sigma recognition for top academic performance. Ms. Olayinka currently contributes to the IEEE MGA SAC Industry Relations Subcommittee, helping develop digital platforms that bridge academia and global tech stakeholders.

She is the creator of frameworks on agentic AI in cybersecurity and anomaly detection in IoT networks, and her Google Scholar and ResearchGate profiles reflect a growing research footprint in applied AI. She is currently preparing to begin a Ph.D. in cybersecurity and AI policy, with the goal of advancing trust-centered AI systems for national resilience and global innovation.

**Sumeet Jeswani**[2] received the B.E. degree in Electronics and Telecommunications from the University of Mumbai, India, in 2013, and the M.S. degree in Telecommunications engineering from the University of Maryland, College Park, MD, USA, in 2018.

He was a Software Engineer with Capgemini, Mumbai, India, from 2013 to 2016, where he worked on cloud security applications and was later promoted to Team Lead. From 2018 to 2021, he was a Solutions Architect and Technical Project Manager with Akamai Technologies, Denver, CO, USA, where he led large-scale enterprise application and network security initiatives for Fortune 500 clients in the financial services industry. Since 2021, he has been a Solutions Consultant with Google, Boulder, CO, USA, where he leads secure cloud transformation projects and GenAI security risk assessments. His work includes architecting secure end-to-end application stacks, implementing zero-trust security controls, and advancing the secure adoption of enterprise-grade AI systems.

Mr. Jeswani is an active member of the IEEE and actively contributes to academia and the broader technical community. He has mentored undergraduate students at the University of Colorado, Boulder, and served as a guest speaker at the University of Colorado, Denver on Responsible AI and Cybersecurity. He has judged AI security hackathons and regularly contributes to academic workshops and industry forums on emerging topics in cloud and AI security."

**Divine Iloh**[3] received the B.S. degree in Public Administration from the University of Nigeria, and the M.S. degree in Business Information Systems and Analytics from the University of Arkansas at Little Rock. He is a researcher and data analyst specializing in artificial intelligence, machine learning, educational technology, and data-driven systems.

He is the co-founder of SabiScholar, an AI-driven EdTech platform that delivers personalized learning to students in emerging markets. He has contributed to peer-reviewed research in AI, cybersecurity, and IoT applications, with an emphasis on agentic systems, anomaly detection, and context-aware automation. His work focuses on developing intelligent systems and AI-powered tools that enhance learning, decision-making, and digital resilience. His current research interests include AI for social impact, intelligent learning systems, IoT-based education solutions, and the application of machine learning in real-world environments.

Mr. Iloh is an active member of IEEE and supports interdisciplinary efforts that bridge AI, education, and equitable access to technology.